\documentclass[sigconf]{acmart}

\usepackage{subfigure}
\usepackage{graphicx}
\usepackage{enumitem}
\usepackage{multirow}

\AtBeginDocument{%
\providecommand\BibTeX{{%
\normalfont B\kern-0.5em{\scshape i\kern-0.25em b}\kern-0.8em\TeX}}}

\copyrightyear{2020} 
\acmYear{2020} 
\setcopyright{acmcopyright}\acmConference[MM '20]{Proceedings of the 28th ACM International Conference on Multimedia}{October 12--16, 2020}{Seattle, WA, USA}
\acmBooktitle{Proceedings of the 28th ACM International Conference on Multimedia (MM '20), October 12--16, 2020, Seattle, WA, USA}
\acmPrice{15.00}
\acmDOI{10.1145/3394171.3413560}
\acmISBN{978-1-4503-7988-5/20/10}

\acmSubmissionID{2390}

\begin{document}
{\onecolumn
\noindent \vspace{1cm}

\noindent \textbf{\huge{Hybrid Dynamic-static Context-aware Attention Network for Action Assessment in Long Videos}}

\vspace{2cm}

\noindent {\LARGE{Ling-An Zeng, Fa-Ting Hong, Wei-Shi Zheng, Qi-Zhi Yu, Wei Zeng, Yao-Wei Wang, and Jian-Huang Lai}}

\vspace{2cm}



\noindent For reference of this work, please cite:

\vspace{1cm}
\noindent Ling-An Zeng, Fa-Ting Hong, Wei-Shi Zheng, Qi-Zhi Yu, Wei Zeng, Yao-Wei Wang, and Jian-Huang Lai.
Hybrid Dynamic-static Context-aware Attention Network for Action Assessment in Long Videos \emph{Proc. of ACM International Conference on Multimedia (ACM MM),} 2020.  

\vspace{1cm}
\noindent Bib:\\
\noindent
@inproceedings\{zeng2020hybrid,\\
\ \ \   title=\{Hybrid Dynamic-static Context-aware Attention Network for Action Assessment in Long Videos\},\\
\ \ \  author=\{Ling-An Zeng, Fa-Ting Hong, Wei-Shi Zheng, Qi-Zhi Yu, Wei Zeng, Yao-Wei Wang, and Jian-Huang Lai\},\\
\ \ \  booktitle=\{ACM International Conference on Multimedia\},\\
\ \ \  year=\{2020\}\\
\}
}

\title{Hybrid Dynamic-static Context-aware Attention Network for Action Assessment in Long Videos}


\author{Ling-An Zeng$^1$, Fa-Ting Hong$^1$, Wei-Shi Zheng$^{1}$  }
\authornote{Corresponding author}

\author{Qi-Zhi Yu$^2$, Wei Zeng$^3$, Yao-Wei Wang$^4$, Jian-Huang Lai$^1$}
        
\affiliation{
  \institution{{$^1$} Sun Yat-sen University, China \quad
  {$^2$} Zhejiang Laboratory \quad
  {$^3$} Peking University, China \quad
  {$^4$} PengCheng Laboratory \\
  \{zenglan3, hongft3\}@mail2.sysu.edu.cn \quad
  wszheng@ieee.org \quad
  yuqz@zhejianglab.com \quad
  }
}
\affiliation{
  \institution{
  weizeng@pku.edu.cn \quad
  wangyw@pcl.ac.cn \quad
  stsljh@mail.sysu.edu.cn
  }
}
 







\renewcommand{\shortauthors}{Zeng et al.}
\renewcommand{\authors}{Ling-An Zeng, Fa-Ting Hong, Wei-Shi Zheng, Qi-Zhi Yu, Wei Zeng, Yao-Wei Wang and Jian-Huang Lai }

\begin{abstract}
The objective of action quality assessment is to score sports videos. However, most existing works focus only on video dynamic information (i.e., motion information) but ignore the specific postures that an athlete is performing in a video, which is important for action assessment in long videos. In this work, we present a novel hybrid dynAmic-static Context-aware attenTION NETwork (ACTION-NET) for action assessment in long videos. To learn more discriminative representations for videos, we not only learn the video dynamic information but also focus on the static postures of the detected athletes in specific frames, which represent the action quality at certain moments, along with the help of the proposed hybrid dynamic-static architecture. Moreover, we leverage a context-aware attention module consisting of a temporal instance-wise graph convolutional network unit and an attention unit for both streams to extract more robust stream features, where the former is for exploring the relations between instances and the latter for assigning a proper weight to each instance. Finally, we combine the features of the two streams to regress the final video score, supervised by ground-truth scores given by experts. Additionally, we have collected and annotated the new Rhythmic Gymnastics dataset, which contains videos of four different types of gymnastics routines, for evaluation of action quality assessment in long videos. Extensive experimental results validate the efficacy of our proposed method, which outperforms related approaches. The codes and dataset are available at \url{https://github.com/lingan1996/ACTION-NET}.
\end{abstract}
\vspace{-0.2cm}


\vspace{-0.2cm}
\vspace{-0.2cm}

\maketitle

\vspace{-0.2cm}

{\small\textbf{ACM Reference Format:} \\
Ling-An Zeng, Fa-Ting Hong, Wei-Shi Zheng, Qi-Zhi Yu, WeiZeng,  Yao-Wei  Wang and Jian-Huang Lai.  2020.  Hybrid  Dynamic-static Context-aware Attention Network for Action Assessment in Long Videos. In \emph{Proceedings of the 28th ACM International Conference on Multimedia (MM ’20), October 12–16, 2020, Seattle, WA, USA}. ACM, New York, NY, USA, 9 pages. https://doi.org/10.1145/1122445.1122456}


\section{Introduction}

In certain types of sports competitions, athletes' scores are given by experts, who have trained in the corresponding field for many years, based on the movements made by the athletes being scored. However, without onsite evaluations by such experts, athletes cannot obtain rapid feedback to improve their performance. It is therefore natural to ask whether a computer vision model can be built to automatically assess the quality of the actions performed by athletes. It is already known that such action quality assessment can be beneficial in other fields, such as medical treatment and the teaching of experimental methods in a scholastic setting.

\begin{figure}
    \centering
    \includegraphics[width=\linewidth]{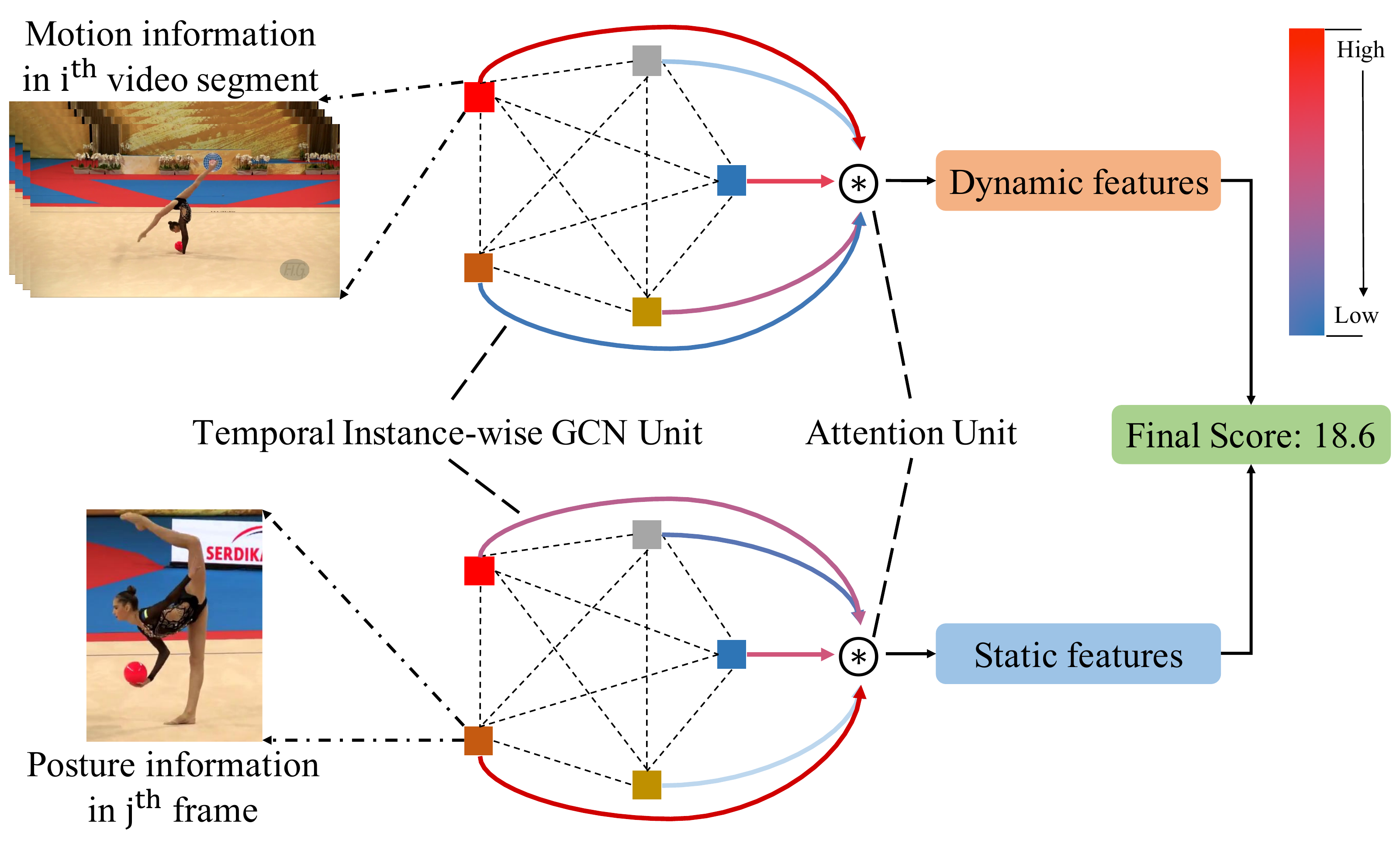}
\vspace{-0.3cm}
\caption{Action quality assessment by ACTION-NET. ACTION-NET takes both motion information and posture information as inputs. Here, we utilize posture information to effectively supplement the motion information. We also adopt an attention mechanism to exploit the important part in the video.
}
    \label{Introduction}
\vspace{-0.5cm}
\end{figure}

Recently, action quality assessment has received increasing interest in the computer vision community \cite{pirsiavash2014assessing,parmar2017learning,zia2018video,li2018scoringnet,li2018end,doughty2018s,doughty2019pros,xu2019learning,pan2019action}.
However, most existing works \cite{parmar2017learning,li2018scoringnet,li2018end,doughty2018s,doughty2019pros,xu2019learning} focus only on \emph{dynamic information} (i.e., motion information), which mainly reflects the category of an action, but ignore \emph{static information}, which reflects the quality of an action at a certain moment. For instance, an insufficient angle of leg lift will result in a substandard action and penalty points for a leg-lifting action. Although an incorrect angle of leg lift may be easily missed on dynamic information, it can be clearly identified from a specific video frame. 
Moreover, for long videos, existing works \cite{pirsiavash2014assessing, parmar2017learning, zia2018video,doughty2018s,pan2019action} simply treat the importance of each part of a video as the same. However, actions performed at different times may have markedly different effects on the final score issued by experts, meaning that the different parts of a video are not of equal importance. For example, in artistic gymnastics, the highlights of a routine should be more important than other parts. 
Consequently, these works have poor performance for action assessment in long videos.

To address the first problem mentioned above, we propose a \emph{ hybrid dynamic-static architecture} for learning video representations, as depicted in Figure \ref{Introduction}. There are two streams in the proposed architecture: a \emph{dynamic stream} for extracting motion information from video, which takes video segments as inputs, and a \emph{static stream} for exploring the postures of detected athletes in specific frames. The same network structure is used for both streams, but the parameters are not shared. By fusing these two separate streams, we can model more discriminative features to represent the actions. Additionally, rather than treating each segment in a video equally, we apply a \emph{context-aware attention module} to each stream to generate more informative \emph{dynamic/static features} before they are concatenated to regress the final score for the video. More specifically, the context-aware attention module consists of a temporal instance-wise graph convolutional network (GCN) unit and an attention unit, the former for exploring the relations between instances and the latter for assigning a proper weight to each instance, for both streams to extract more robust stream features. Subsequently, we aggregate the fused local-context features of all segments/frames (or motions/postures) weighted with the weights estimated by the attention unit to generate the final dynamic/static features.

To support research on action quality assessment for long videos, we have constructed a new Rhythmic Gymnastics dataset. The Rhythmic Gymnastics dataset contains videos of four different types of gymnastics routines: ball, clubs, hoop and ribbon. Each type of routine has 250 associated videos, and the length of each video is approximately 1 min 35 s. We chose high-standard international competition videos, including videos from the 36th and 37th International Artistic Gymnastics Competitions, to construct the dataset. We have edited out the irrelevant parts of the original videos (such as replay shots and athlete warmups). We have annotated each video with three scores (a difficulty score, an execution score and a total score), which were given by the referee in accordance with the official scoring system. This dataset will be released soon.

In summary, we propose a hybrid dynAmic-statiC conText-aware attentION NETwork (ACTION-NET) to predict athlete scores from long videos. To the best of our knowledge, ACTION-NET is the first such technology to incorporate both video motion information and posture information for detected people in static frames for action quality assessment. Experiments conducted on the public MIT-Skating dataset and our Rhythmic Gymnastics dataset clearly show that our method achieves state-of-the-art results. The codes and dataset are available at \url{https://github.com/lingan1996/ACTION-NET}.

\section{Related Work}

\noindent{\textbf{Action Quality Assessment.}}
The challenging problem of action quality assessment in videos has been extensively explored in previous works \cite{pirsiavash2014assessing,li2018scoringnet,li2018end,xu2019learning,parmar2017learning,bertasius2017baller,zia2018video,doughty2018s,doughty2019pros,parmar2016measuring,zia2015automated}. The existing methods can be divided into two strategies based on the information used as the input---namely, pose-based methods and vision-based methods. Pose-based methods \cite{pirsiavash2014assessing,pan2019action} analyze human pose information to assess action quality. Pirsiavash et al. \cite{pirsiavash2014assessing} used the DCT features of joint trajectories as the input for SVR to predict final scores supervised by ground-truth scores given by referees. Pan et al. \cite{pan2019action} considered the relations among joints and proposed two modules---namely, a joint commonality module and a joint difference module---to analyze joint motion. However, due to the atypical body postures involved, it is difficult to estimate the human poses executed during sports such as diving. Moreover, joint motion alone cannot reflect appearance information, which serves as the basis for points scored in gymnastics, or background information, such as the splash size in diving. By contrast, vision-based methods \cite{li2018scoringnet,li2018end,xu2019learning,parmar2017learning,doughty2019pros} assess the quality of actions based on visual features extracted from videos by 3D convolutional neural networks (CNNs). For instance, to model more informative features from specific fragments, key fragment segmentation \cite{li2018scoringnet} has been proposed to obtain key fragments while discarding irrelevant fragments. In addition, Li et al. \cite{li2019manipulation} proposed a novel recurrent neural network (RNN)-based spatial attention model based on the attention mechanism used by humans when assessing videos. However, because these video feature extraction methods are designed for action classification, the dynamic information they extract mainly reflects the action category rather than the action quality.

For long videos, actions performed at different times may have markedly different effects on the final score issued by experts. However, most existing works \cite{pirsiavash2014assessing, parmar2017learning,pan2019action} simply treat the importance of each part from a video equally, so the important part may not be explored effectively in such a way. To assess the quality of actions in long videos, several different methods have been proposed \cite{bertasius2017baller,doughty2019pros,xu2019learning}. Bertasius et al. \cite{bertasius2017baller} used a convolutional long short-term memory (LSTM) network to detect atomic basketball events, such as shooting or passing the ball during a basketball game, from first-person videos. However, because that method is a strongly task-related method, it is difficult to generalize. Xu et al. \cite{xu2019learning} used a self-attentive LSTM network and a multiscale skip LSTM network to learn local (technical movements) and global (athlete performance) scores, respectively. Doughty et al. \cite{doughty2019pros} proposed a rank-aware temporal attention mechanism to attend to the skill-relevant parts of a video. However, these works did not consider the temporal relations between adjacent parts of a video.

In contrast to previous work, we propose a hybrid dynamic-static architecture for learning both dynamic information and static information related to specific moments. We argue that in sports, static information is also important for exploring the postures of the athletes detected in specific frames to determine whether those postures are correct. Additionally, we propose a context-aware attention module to aggregate all video segments/frames (motions/postures) to produce dynamic/static features.

\vspace{5pt}

\noindent{\textbf{Video Understanding.}}
Video understanding has been studied for a long time in the field of computer vision and provides fundamental tools for action quality assessment. A number of deep neural network architectures for action classification have been proposed \cite{simonyan2014two, wang2016temporal, tran2015learning, carreira2017quo, yue2015beyond, yang2018making}, including two-stream networks with multiple modalities \cite{simonyan2014two,wang2016temporal}, 3D CNNs for extracting spatial and temporal features \cite{tran2015learning, carreira2017quo}, and RNNs for capturing temporal relations in variable-length videos \cite{yue2015beyond,yang2018making}. In addition to action classification, many works \cite{saha2016deep,zolfaghari2017chained,alwando2018cnn,peng2016multi} have studied temporal action localization, with the aim of generating sequences of bounding boxes related to the locations of the actors.

In this paper, we employ the I3D architecture \cite{carreira2017quo} to obtain the basic video features from the video segments. Because the video features extracted by I3D are not designed for action quality assessment and mainly reflect the action category, we especially add a static stream, from which features are extracted using the ResNet architecture \cite{he2016deep} to explore the postures of detected athletes in specific frames.

\section{Method}

\begin{figure*}[ht]
 \centering
 \includegraphics[width=17cm, height=8.5cm]{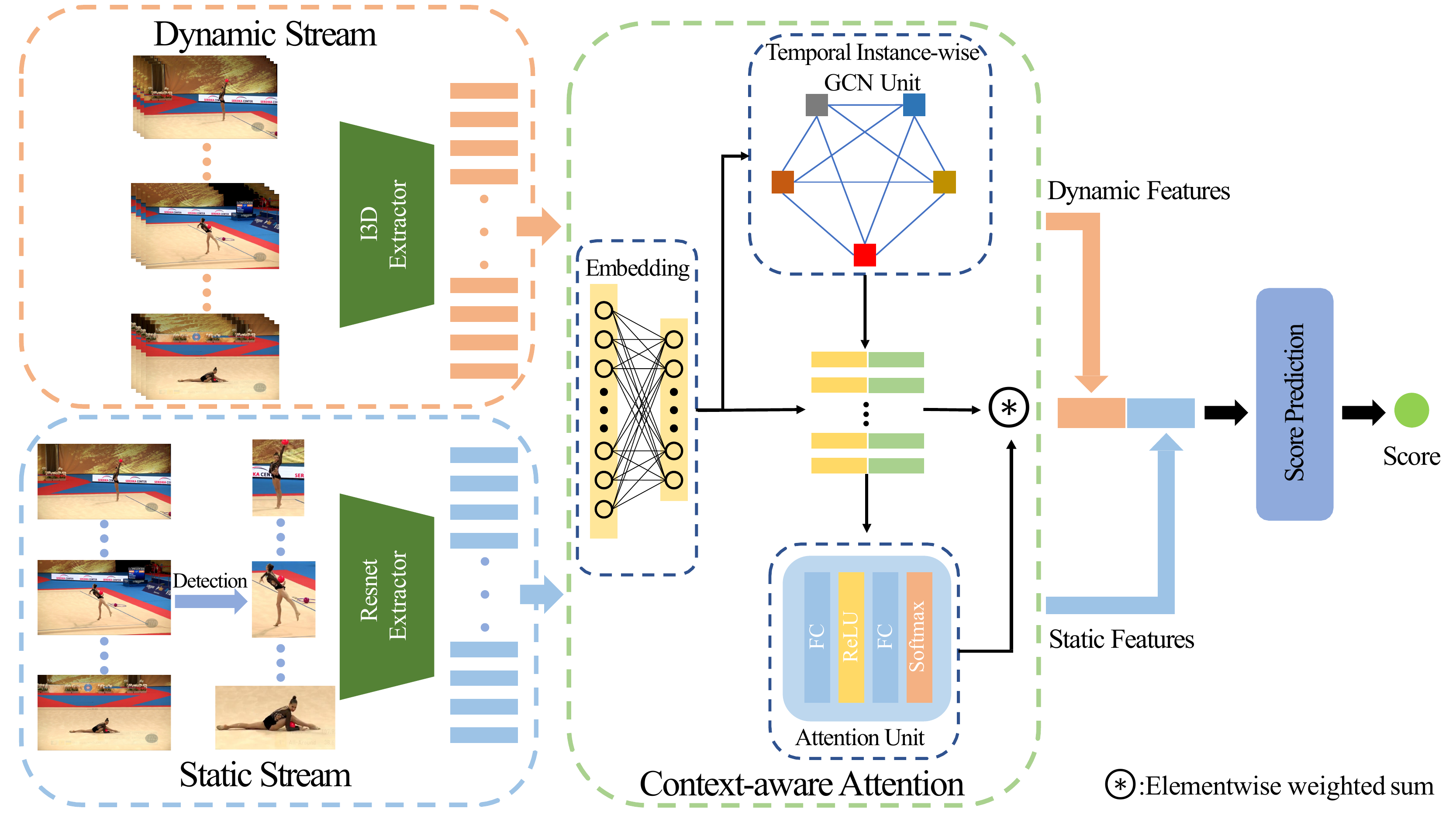}
\vspace{-0.3cm}
\caption{\textbf{The pipeline of our ACTION-NET.} Based on the dynamic stream, motion information and background information are extracted from video segments. The static stream provides spatial information about the postures and appearance of detected athletes in specific frames. Both streams are fed into ACTION-NET branches with identical structures, but these branches do not share parameters. The proposed context-aware attention module learns the relations between all segments/frames and generates dynamic/static features by aggregating the fused local-context features of all segments/frames. Finally, we concatenate the features from both streams and feed them into the score prediction module.}
 \label{pipeline}
\Description{pipeline of model}
\vspace{-0.3cm}
\end{figure*}

In this work, we develop ACTION-NET for action quality assessment. By leveraging two streams---i.e., a dynamic stream and a static stream---ACTION-NET generates more informative representations for videos by exploiting video motion information and the posture information of detected athletes in specific frames, respectively. Moreover, we propose a context-aware attention module to aggregate all video segments/frames to produce dynamic/static features in each stream. More details of the proposed network can be seen in Figure \ref{pipeline}. In this section, we present detailed descriptions of the steps of action quality assessment using ACTION-NET. We introduce the proposed context-aware attention module in section \ref{sec:attention} after first introducing the dynamic and static streams of the hybrid dynamic-static architecture in section \ref{sec:archi}.

\vspace{-0.1cm}
\subsection{Hybrid Dynamic-Static Architecture} \label{sec:archi}

\begin{figure}
 \centering
 \includegraphics[width=\linewidth, height=3cm]{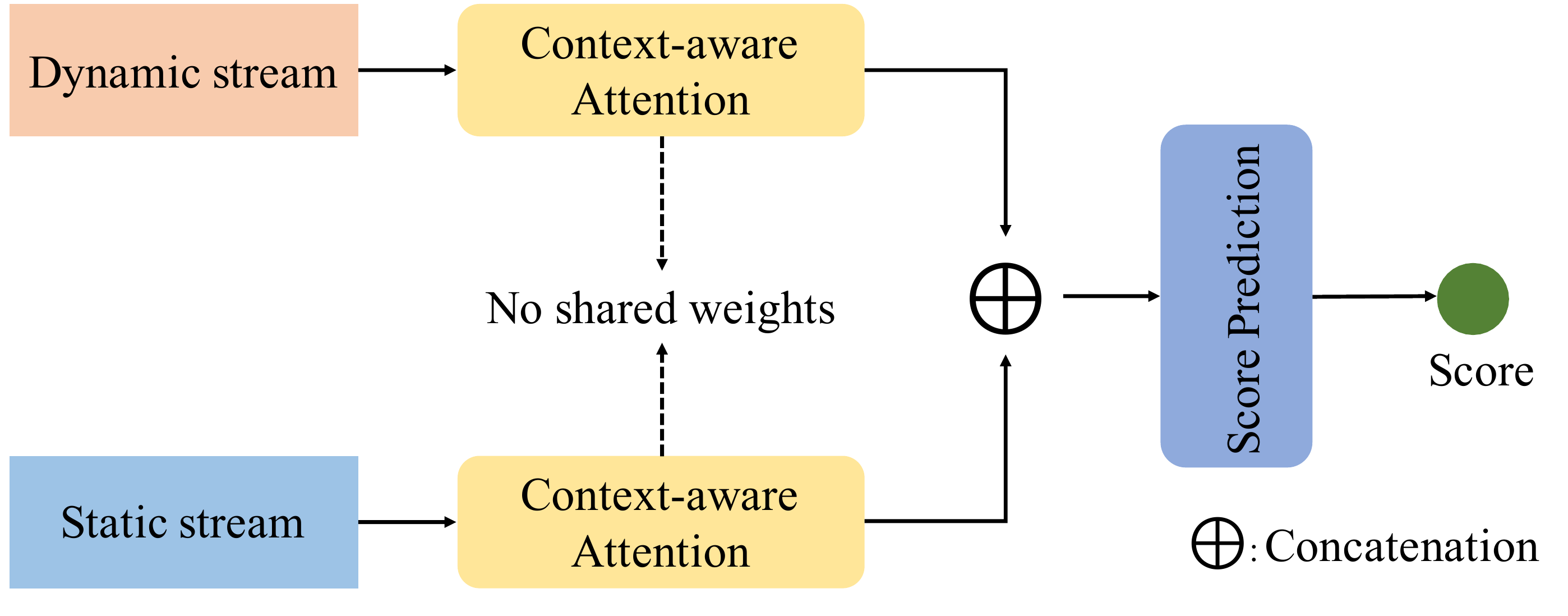}
\vspace{-0.3cm}
\caption{Hybrid dynamic-static architecture. Two streams have different purposes but identical streams. Here, the dynamic stream mainly extracts the motion information of video, and the static stream extracts the posture information of specific frames.
}
 \label{ Dynamic-static hybrid Architecture}
\Description{ Dynamic-static hybrid Architecture}
\vspace{-0.5cm}
\end{figure}

Most existing works \cite{parmar2017learning,li2018scoringnet,li2018end,doughty2018s,doughty2019pros,xu2019learning} mainly focus on the dynamic temporal information contained in videos and thus ignore incorrect postures at specific moments during an athlete's routine because of the fleeting nature of these postures. To address this problem, we propose a hybrid dynamic-static architecture for learning both dynamic information and posture information at specific moments, as depicted in Figure \ref{ Dynamic-static hybrid Architecture}. Specifically, the network branches corresponding to the two streams have the same structure but do not share parameters.

\vspace{5pt}

\noindent\textbf{Dynamic Stream.} We leverage a stream of segments from the same video as the input to explore the motion information of the video. Instead of operating on the basis of optical flow, our dynamic stream structure utilizes a sequence of video segments sampled from the raw video. For the dynamic stream, we first feed the segments into an I3D network pretrained on the Kinetics dataset \cite{carreira2017quo} to extract 1024-dimensional segment features, which are then passed to the context-aware attention module. In the context-aware attention module, we compute the context-related features for each segment by a temporal instance-wise graph convolutional network unit (TCG-U), which is leveraged to model the relationship between 
each segment and the corresponding global information (with all segment features as inputs).
An attention unit (ATT-U) then takes the fused local-context features obtained by concatenating the segment features and the context-related feature as inputs to generate an attention weight for each segment. Finally, we aggregate the fused local-context features of all segments weighted with the attention weights estimated by the ATT-U to generate dynamic features.

\vspace{5pt}

\noindent\textbf{Static Stream.} As mentioned before, the video segments can yield dynamic information, but slightly incorrect postures can easily be ignored because of their fleeting nature. To address this challenge, we adopt a static stream, for which the network structure is the same as that for the dynamic stream, to provide supplemental posture and appearance information of the detected athletes. For the static stream, we sample frames from raw videos to explore the posture information. Since the athletes in sports videos often occupy only a small part of the image, it is difficult to extract high-quality spatial features of an athlete when the entire image is taken as the input. Therefore, we first perform human detection on the sampled frames and then extract features from the detected people. In this way, we can obtain high-quality spatial information on the exhibited postures. Subsequently, we feed the extracted posture features from each frame into the context-aware attention module for the static stream, which is the same as that for the dynamic stream, to generate static features.

\vspace{-0.1cm}
\subsection{Context-Aware Attention Module} \label{sec:attention}

As described in section \ref{sec:archi}, a context-aware attention module is applied to both streams to aggregate all segment/frame (motion/posture) features to produce the features of the corresponding stream (i.e., dynamic features and static features). Our proposed context-aware attention module consists of two units: the TCG-U for modeling the relationships between the segments/frames and the ATT-U for estimating the attention weights of these segments/frames.

\vspace{5pt}

\noindent\textbf{Temporal Clipwise GCN Unit (TCG-U).} After obtaining the features $\{\mathbf{f}_I^i\}_{i=1}^N$ of each instance (here, we refer to both segments and frames as instances for convenience of description), we use a GCN to output the context-related features $\{\mathbf{f}_H^i\}_{i=1}^N$ for each instance by aggregating all instance features. To iteratively learn the relationships between all instances, we construct a graph $\mathcal{G}$ with all instances as vertices. We then adopt an exponential kernel \cite{zhong2019graph} to compute the adjacency matrix $A \in \mathbb{R}^{N\times N}$ for graph $\mathcal{G}$:
\begin{equation}
A_{(i,j)} = exp^{\frac{-||\mathbf{f}_I^i-\mathbf{f}_I^j||}{K}},
\end{equation}
where $K$ is a positive hyperparameter used to adjust the scale of the distance between two vertices, and element $A_{(i,j)}$ of the adjacency matrix $A$ represents the temporal relationship between the $i^{th}$ and $j^{th}$ instances. In our experiment, $K$ is set to 1.

It is known that the adjacent instances in a graph exchange information through iterative graph-Laplacian operations.
To maintain the original distribution of the matrix $A$

in the process of information transmission, we normalize the adjacency matrix, following Kipf and Welling \cite{kipf2016semi}:
\begin{equation}
\widehat{A} = \widetilde{D}^{-\frac{1}{2}} \widetilde{A} \widetilde{D}^{-\frac{1}{2}},
\end{equation}
where the matrix $\widetilde{A}$ is defined as $\widetilde{A} = A + I $, with $I \in \mathbb{R}^{N \times N} $ being the identity matrix, and $\widetilde{D}_{(i,i)} = \sum_{i,j} \widetilde{A}_{i,j}$ is the degree matrix of the adjacency matrix variant $\widetilde{A}$.
To learn more context knowledge from the graph $\mathcal{G}$ consisting of all instances, we iteratively update the representation of each vertex by multiple GCN layers to generate more robust context-related features.
\begin{equation}
H^{l+1} = \sigma( \widehat{A} H^{l} W^{l}),
\end{equation}
where $\sigma(\cdot) $ is the $ReLU(\cdot)$ activation function in this work, $W^{l}$ is the trainable weight matrix in the $l^{th}$ layer, and $H^{l}$ is the activation matrix in the $l^{th}$ layer, with $H^{0} =  \{\mathbf{f}_I^i\}_{i=1}^N$. We then use two GCN layers to capture the context information among all instances based on the temporal relations of adjacent instances. Finally, we denote the last $H^{l}$ by $H = \{\mathbf{f}_R^i\}_{i=1}^N$ as the context-related feature for each instance.

\vspace{5pt}

\noindent\textbf{Attention Unit (ATT-U).} After obtaining the context-related features $\{\mathbf{f}_H^i\}_{i=1}^N$, we combine the corresponding instance features to obtain the fused local-context features $\{\mathbf{f}_C^i\}_{i=1}^N$, which contain both instance and global context information and are more representative of each instance in the video:

\begin{equation}
\mathbf{f}_C^i = concatenate(\mathbf{f}_I^i,\mathbf{f}_R^i).
\end{equation}

Here, we concatenate the instance features $\mathbf{f}_I^i$ and the context-related features $\mathbf{f}_R^i$ to obtain the fused local-context features $\mathbf{f}_C^i$.

We then calculate the weighted sum of all fused local-context features using the attention weights $\{\varepsilon_i\}_{i=1}^N$ estimated by our simple ATT-U, which consists of two fully connected layers and corresponding activation function with fused local-context features as inputs, to produce the corresponding stream features $\mathbf{f}_{D/S}$ (i.e., dynamic features $\mathbf{f}_{D}$ or static features $\mathbf{f}_{S}$):
\begin{equation}
\begin{aligned}
\mathbf{f}_{D/S} &= \sum_{i}{\varepsilon_i\mathbf{f}_{C}^{i}},\\
\varepsilon_{i} &= \alpha(\mathbf{f}_{C}^{i}).
\end{aligned}
\end{equation}
where $\alpha(\cdot)$ represents the ATT-U. In this way, we utilize all instances in a video to construct fused local-context features that capture not only the corresponding instance attributes but also the context information.

\vspace{-0.1cm}
\subsection{Combination of Dynamic and Static Streams}
Finally, we obtain the dynamic and static features from their respective streams. We concatenate these two types of features and regress the final score for the input video as follows:
\begin{equation} \label{eq:predict}
s = f_r([\mathbf{f}_{D}; \mathbf{f}_{S}]),
\end{equation}
where $f_r(\cdot)$ represents two fully connected layers with the sigmoid activation function. The sigmoid activation function is used to normalize the estimated score to the range of $[0, 1]$.

\vspace{-0.1cm}
\section{Rhythmic Gymnastics Dataset}

In addition, we construct a new dataset for action quality assessment---i.e., the Rhythmic Gymnastics dataset---which contains videos of four types of gymnastics routines: ball, clubs, hoop and ribbon. We collected these videos from the internet by searching for specific topics. All videos in the Rhythmic Gymnastics dataset are of high quality and were downloaded from the official account on YouTube.

The Rhythmic Gymnastics dataset consists of 1000 videos spanning all types of routines---\i.e., ball, clubs, hoop and ribbon---with 250 videos per routine type. More specifically, these videos were obtained from high-standard international competitions---i.e., the 36th and 37th International Artistic Gymnastics Competitions. Because the scoring criteria for rhythmic gymnastics changed significantly after the 35th International Rhythmic Gymnastics Competition, we did not select any videos from before the scoring change to ensure the uniformity of the scoring criteria. 

\vspace{5pt}

\noindent{\textbf{Preprocessing.}} We first collected approximately 37 h of videos and removed abnormal videos, such as those in which the athlete retired for atypical reasons. We edited out the irrelevant parts of each original video (such as bowing to the audience and warming up). We preserved only the duration of each video from the moment of the beginning pose to the moment of the ending pose. The length of each video is approximately 1 min 35 s, corresponding to approximately 2375 frames at a frame rate of 25 frames per second.

\vspace{5pt}

\noindent{\textbf{Score Annotation.}} We annotated each video with three scores (a difficulty score, an execution score and a total score), which were given by the referee in accordance with the scoring system. The final score is the sum of the difficulty score and the execution score, with deductions for any penalties incurred. The difficulty score consists of subscores for body difficulties, dynamic elements with rotation (commonly known as risks), and apparatus difficulties. The execution score represents the degree to which the gymnast performs with aesthetic and technical perfection. Finally, for each event, we randomly split the dataset into 200 training videos and 50 testing videos.

\vspace{-0.1cm}

\section{Experiments}

In this section, we first describe the implementation details of our ACTION-NET model and then present the experimental results obtained on two datasets---i.e., the public MIT-Skating dataset and our Rhythmic Gymnastics dataset---and compare them with the results of several baselines. Finally, we conduct an ablation study to analyze the contributions of the hybrid dynamic-static architecture and the context-aware attention module as well as the necessity of human detection on the static stream.

\vspace{-0.1cm}

\subsection{Datasets and Evaluation Metric}

\textbf{Datasets.} We evaluated our ACTION-NET model on both the MIT-Skating dataset \cite{pirsiavash2014assessing} and our newly constructed Rhythmic Gymnastics dataset. The MIT-Skating dataset contains 150 figure skating videos, each of which is approximately 2 min 55 s. Each one contains an average of 4200 frames. Each video is labeled with three scores---i.e., a total element score, a total program component score and a final score; the final score, which ranges from 0 (worst) to 100 (best), is the sum of the total element score and the total program score. We used 100 videos for training and the remaining 50 videos for testing. \cite{parmar2016measuring,parmar2017learning} repeated the experiment 200 times with different random data splits and averaged the results. Due to the heavy computing cost, we repeated the experiment on five random data splits instead. On the Rhythmic Gymnastics dataset, for each type of routine, we used 200 videos for training and the remaining 50 videos for testing.

\vspace{5pt}

\noindent{\textbf{Evaluation Metric.}}
Following existing works \cite{pirsiavash2014assessing,li2018end,xu2019learning}, we used the standard evaluation metric known as Spearman's rank correlation coefficient $\rho$, which represents the strength of the relation between two series of data. Its value ranges from -1 to 1, and it is computed as follows:
\begin{equation}
\rho = \frac 
{\sum _{i} {(x_i - \overline{x}) (y_i - \overline{y}) }} 
{\sqrt{ \sum _{i} {(x_i - \overline{x})^2}     \sum _{i} {(y_i - \overline{y})^2} }},
\end{equation}
where $x$ and $y$ represent the rankings of the two series. The higher the value of $\rho$, the higher the rank correlation between the predicted and ground-truth scores.

\vspace{-0.1cm}
\subsection{Implementation Details}

\noindent{\textbf{Data Preprocessing.}}
Similar to previous works \cite{parmar2017learning,xu2019learning}, our method is composed of two stages: feature extraction and score prediction. We first extracted features from videos and then used our network to predict scores. For the dynamic stream, we sampled $5$ frames per second on average, and each segment contained 16 sampled frames. All frames were rescaled to have a shortest side length of 256 and were then center cropped to $224 \times 224$. We then extracted 1024-dimensional segment features from the \emph{avg\_pool} layer of I3D pretrained on Kinetics \cite{carreira2017quo}. For the static stream, we sampled $1$ frame per second and cropped the region with the detected athlete in the sampled frame using YOLOv3 \cite{redmon2018yolov3} pretrained on COCO \cite{lin2014microsoft}. To filter out the action-irrelevant audience, we chose only the largest human bounding box in each frame and discarded all frames in which no athlete was detected. All cropped image regions were rescaled to $224 \times 224$. We then extracted the 2048-dimensional features from the \emph{avg\_pool} layer of ResNet \cite{he2016deep} pretrained on ImageNet \cite{russakovsky2015imagenet}.

\vspace{5pt}

\noindent{\textbf{Experimental Settings.}}
In the context-aware attention module, two fully connected layers with ReLU activation first embedded the input features into a low-dimensional space. For the dynamic stream, the dimensions of these two layers were $1024 \times 512$ and $512 \times 256$. For the static stream, their dimensions were $2048 \times 1024$ and $1024 \times 256$. Two GCN layers were used for the TCG-U, both with dimensions of $ 256 \times 256$. We used two fully connected layers as $f_r$ in Eq. \ref{eq:predict} to predict the final score; the first fully connected layer had dimensions of $1024 \times 128$ (followed by ReLU activation and dropout = 0.5), and the second layer had dimensions of $128 \times 1$ with sigmoid activation. To ensure stable training, we adopted different learning rates for the context-aware attention module (lr = 0.01) and the score prediction module (lr = 0.05). We used minibatch stochastic gradient descent (SGD) to train the network, with a momentum of 0.9 and a weight decay of $ 10^{-4}$. The batch size was 32 for the Rhythmic Gymnastics dataset and 16 for the MIT-Skating dataset. The number of training epochs was 200 for the MIT-Skating dataset, and for better convergence, we set different training epochs on four types of gymnastics routines (400/300/500/300 for ball/clubs/hoop/ribbon). For the Rhythmic Gymnastics dataset, we applied stepwise decay in the last 100 and last 50 epochs with a decay rate of 0.1. For the MIT-Skating dataset, we applied stepwise decay after 150 and 180 epochs with a decay rate of 0.1. Remarkably, although our model is a hybrid dynamic-static architecture, our model is small, with only 3.54 M parameters.

\vspace{5pt}

\noindent{\textbf{Data Augmentation.}}
For the dynamic stream, the number of video segments per video was approximately 55 on the MIT-Skating dataset and 28 on the Rhythmic Gymnastics dataset. Accordingly, we used 48 video segments per video to train the network on the MIT-Skating dataset and 26 video segments per video on the Rhythmic Gymnastics dataset. For the static stream, the number of cropped frames per video was almost 160 on the MIT-Skating dataset and 80 on the Rhythmic Gymnastics dataset. We used 150 cropped frames per video for training on the MIT-Skating dataset and 80 cropped frames per video on the Rhythmic Gymnastics dataset. Similar to Xu et al. \cite{xu2019learning}, all video segments and cropped frames were augmented by shifting the starting segment and frame.

\vspace{5pt}

\noindent{\textbf{Competitors.}}
We considered several baseline methods for comparison to validate the effectiveness of our proposed method. We first compared our method with those presented in \cite{pirsiavash2014assessing,le2011learning,parmar2017learning,li2018end,xu2019learning}. Because the Rhythmic Gymnastics dataset is a new dataset, to facilitate comparison, we newly evaluated the performance of some previous methods \cite{parmar2017learning, xu2019learning} on this dataset. Following the settings for competitors given in \cite{xu2019learning}, we considered different combinations of the following model components:
\vspace{-0.5cm}
\begin{itemize}
\item Input features: We used either frame-level features or video-segment-level features as the input features. As the frame-level features, we extracted 2048-dimensional features from the \emph{avg\_pool} layer of ResNet \cite{he2016deep}, which is a common feature extractor for images. For the video-segment-level features, we extracted them as described for the dynamic stream.
\item LSTM and Bi-LSTM-based models: Similar to Parmar et al. \cite{parmar2017learning}, we applied an LSTM architecture to generate video-level descriptions. Due to the very long duration of the videos and to ensure fair comparisons, we also tested the use of a bidirectional LSTM (Bi-LSTM) architecture in place of the LSTM architecture. The hidden dimensions of the LSTM/Bi-LSTM layers were 256/128, and to avoid over-fitting, we used one fully connected layer for regression.
\item SVR-based models: We first used either average or maximum pooling for video-level description. Because the effect of the linear kernel was better than that of the radial basis function (RBF) kernel in \cite{xu2019learning}, we applied SVR only with a linear kernel to regress the final scores.
\end{itemize}

\vspace{-0.1cm}
\subsection{Results and Comparisons}

\begin{table}[]
\setlength{\tabcolsep}{1mm}{
\begin{tabular}{|l|c|ccccc|}
\hline
\multicolumn{2}{|c|}{\multirow{2}{*}{}} &
  \multicolumn{1}{c|}{\multirow{2}{*}{MIT-Skating}} &
  \multicolumn{4}{c|}{Rhythmic Gymnastics} \\ \cline{4-7} 
\multicolumn{2}{|c|}{} &
  \multicolumn{1}{c|}{} &
  \multicolumn{1}{c|}{Ball} &
  \multicolumn{1}{c|}{Clubs} &
  \multicolumn{1}{c|}{Hoop} &
  Ribbon \\ \hline
\multicolumn{2}{|c|}{Pose + DCT \cite{parmar2016measuring}}            			& 0.350 & -     & -     & -     & -     \\
\multicolumn{2}{|c|}{ConvISA \cite{le2011learning}}                          			 		& 0.450 & -     & -     & -     & -     \\
\multicolumn{2}{|c|}{C3D + SVR \cite{parmar2017learning} }                       	   & 0.530 & 0.357* & 0.551* & 0.495* & 0.516* \\
\multicolumn{2}{|c|}{Li et al. \cite{li2018end}}     												 	& 0.575 & -    & -     & -     & -     \\
\multicolumn{2}{|c|}{Pan et al. \cite{pan2019action}}&   0.384  & - & - & - &- \\
\multicolumn{2}{|c|}{Xu et al. \cite{xu2019learning} } 									  	 & 0.590 & 0.515* & 0.621* & 0.540* & 0.522* \\ \hline
\multirow{4}{*}{ResNet} & Avg + SVR 	 & 0.545 & 0.279 & 0.589 & 0.602 & 0.395 \\
                       		 				& Max + SVR 	 & 0.512 & 0.169 & 0.452 & 0.376 & 0.404 \\
                        					& LSTM      		 & 0.602 & 0.471 & 0.472 & 0.555 & 0.410 \\
                       						& Bi-LSTM   	   & 0.598 & 0.466 & 0.444 & 0.544 & 0.447 \\ \hline
\multirow{4}{*}{I3D}       & Avg + SVR	 	& 0.531 & 0.403 & 0.579 & 0.610 & 0.456 \\
                        					& Max + SVR 	& 0.442 & 0.445 & 0.465 & 0.386 & 0.191 \\
                        					& LSTM     		 	& 0.472 & 0.327 & 0.583 & 0.546 & 0.381 \\
                        					& Bi-LSTM   	  & 0.587 & 0.355 & 0.653 & 0.557 & 0.368 \\ \hline
\multicolumn{2}{|c|}{ACTION-NET (ours)}          & \textbf{0.615} & \textbf{0.528} & \textbf{0.657} & \textbf{0.708} & \textbf{0.578} \\ \hline
\end{tabular}}
\vspace{0.2cm}
\caption{Results in terms of Spearman's rank correlation coefficient (higher values are better) on the MIT-Skating and Rhythmic Gymnastics datasets.  The results marked with * are those obtained through the reimplementation of \cite{parmar2017learning,xu2019learning} on our Rhythmic Gymnastics dataset. }
\label{Results}
\vspace{-0.6cm}
\end{table}

\begin{table}[]
\setlength{\tabcolsep}{1mm}{
\begin{tabular}{|l|ccccc|}
\hline
\multirow{2}{*}{Method} & \multicolumn{1}{c|}{\multirow{2}{*}{MIT-Skating}} & \multicolumn{4}{c|}{Rhythmic Gymnastics}                                                    \\ \cline{3-6} 
                         & \multicolumn{1}{c|}{}                             & \multicolumn{1}{c|}{Ball} & \multicolumn{1}{c|}{Clubs} & \multicolumn{1}{c|}{Hoop} & Ribbon \\ \hline
DS + CAA       & 0.603 & 0.346 & 0.583 & 0.643 & 0.394 \\
SS + CAA       & 0.575 & 0.443 & 0.581 & 0.686 & 0.540 \\
TS + CAA~(ours) & \textbf{0.615} & \textbf{0.528} & \textbf{0.657} & \textbf{0.708} & \textbf{0.578} \\ \hline
\end{tabular}}
\vspace{0.2cm}
\caption{Ablation study showing the contributions of the dynamic and static streams in our method. 
}
\label{tab:ablation study 1}

\vspace{-0.65cm}
\end{table}

\begin{figure*}[htb]
 \centering
 \includegraphics[width=15cm, height=4cm]{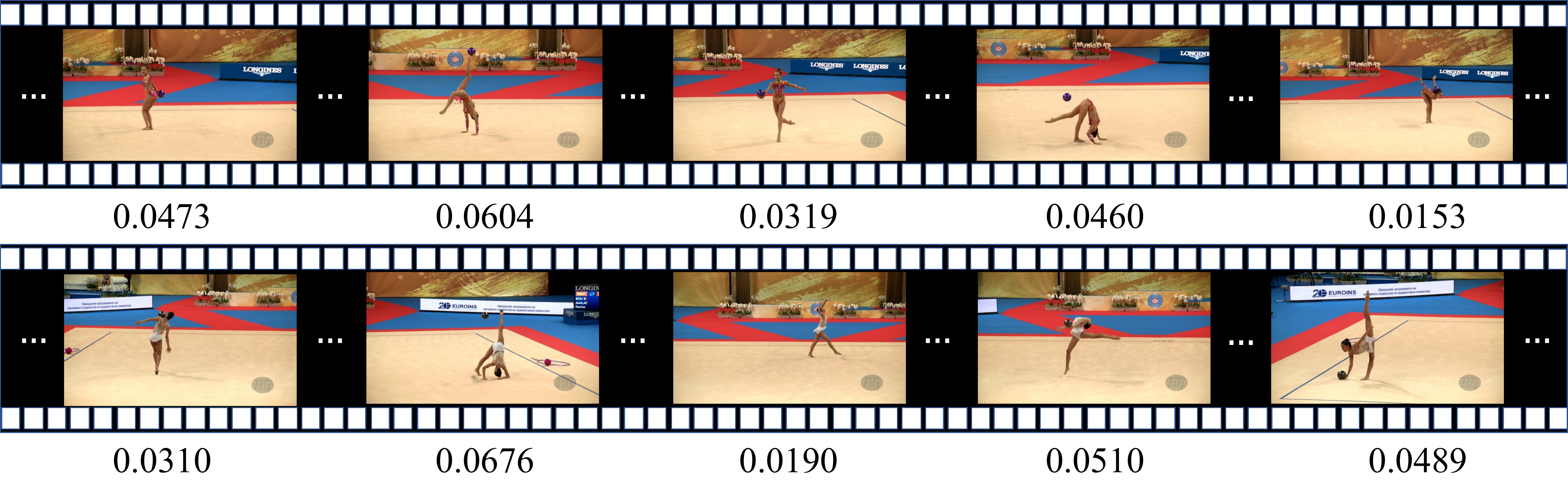}
 \vspace{-0.3cm}
\caption{Visualization of the attention weights generated by our context-aware attention module on two videos. The numbers below the images are the attention weights of the corresponding video segments. 
}
 \label{fig:attention}
\Description{attention}
\vspace{-0.3cm}
\end{figure*}

\begin{figure*}[htb]
 \centering
 \includegraphics[width=17cm]{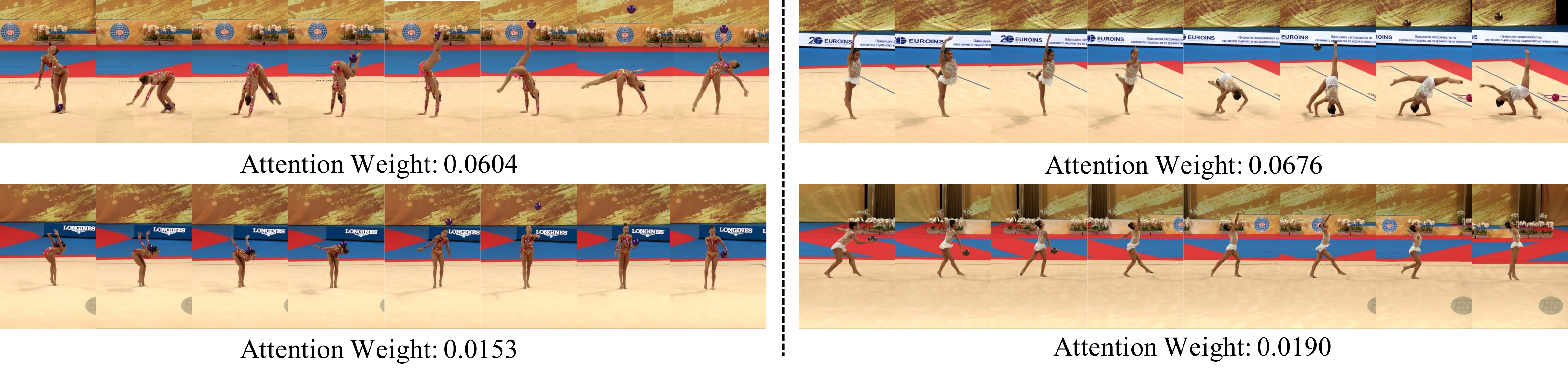}
 \vspace{-0.3cm}
\caption{Visualization of video segments with high and low attention weights from two videos. The first row shows video segments with high attention weights, while the bottom row shows video segments with low attention weights. 
}
 \label{fig:visual}
\Description{visual}
\vspace{-0.3cm}
\end{figure*}

\begin{figure*}[htb]
 \centering
 \includegraphics[width=17cm, height=4.0cm]{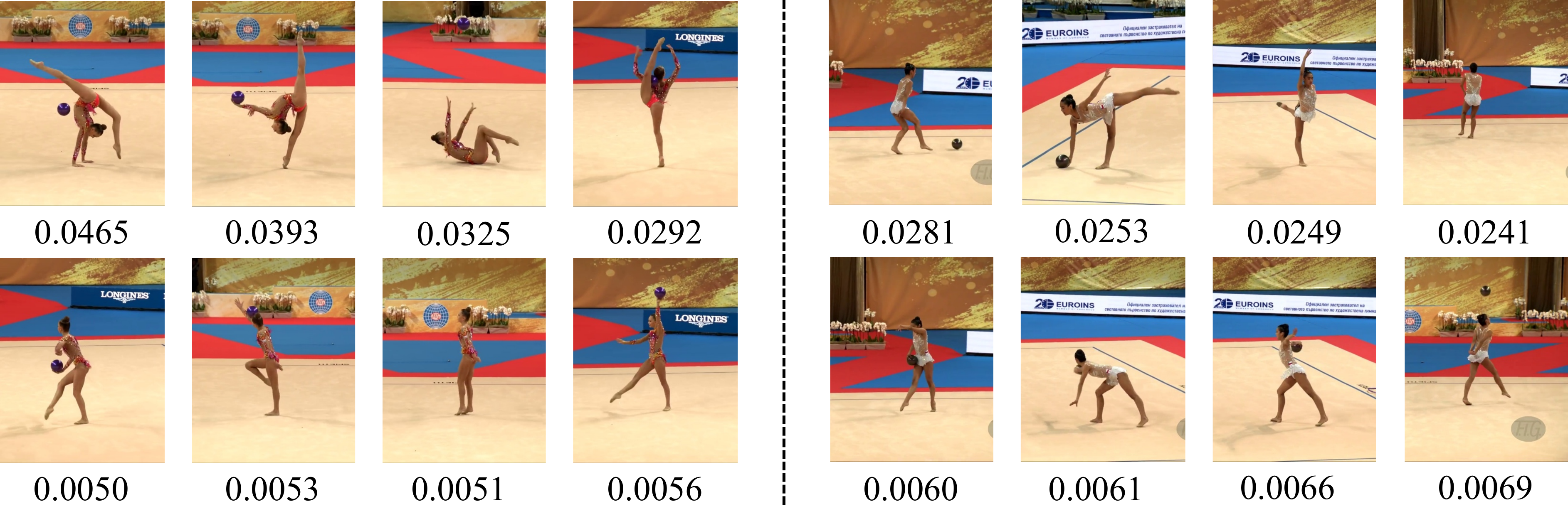}
 \vspace{-0.3cm}
\caption{The top four frames with high or low attention weights in static stream from two videos. The number below each image frame is the attention weight of the corresponding frame. The video frames with high attention show that the gymnast is performing technical movements or making a mistake (e.g., the loss of the apparatus is shown in the $5^{th}$ and $8^{th}$ images in the first row). The video frames with low attention weights show no such important, highly technical postures.}
\vspace{-0.3cm}
 \label{fig:static stream}
\end{figure*}

The experimental results obtained by our method and the compared methods on the MIT-Skating and Rhythmic Gymnastics datasets are presented in Table~\ref{Results}. The results marked with * are those obtained through the reimplementation of \cite{parmar2017learning,xu2019learning} on our Rhythmic Gymnastics dataset. Although Pan et al. \cite{pan2019action} has achieved state-of-the-art results on the AQA-7 dataset \cite{parmar2019action}, which simply contains short videos (102 frames per video on average), their proposal performs poorly on the MIT-Skating dataset \footnote{This result was obtained by Pan through communication. Thanks for their contribution.}. This is because Pan's work simply assumes all parts of a long video are equally important. Our ACTION-NET model achieves the best performance and outperforms the state-of-the-art method presented in \cite{xu2019learning} by 2.5\%. These results of compared methods clearly demonstrate that only utilizing video temporal information is insufficient to model a more comprehensive action assessment. In contrast, the results of our ACTION-NET indicate that the static information helps understand the quality of an action.

Additionally, we further explored different variants of our model, as shown in Table \ref{Results}. By comparing the experimental results of the different variants, we can reach several interesting conclusions. First, in contrast to the results of Xu et al. \cite{xu2019learning}, average pooling is obviously superior to maximum pooling on all datasets.
SVR with average pooling also works well on the Rhythmic Gymnastics dataset but not on MIT-Skating, which is most likely because the videos in MIT-Skating are longer than those in the Rhythmic Gymnastics dataset. Moreover, models using I3D features can produce better prediction results than those using ResNet features extracted from the whole scene.

\vspace{-0.1cm}

\subsection{Model Analysis}

\begin{table}[]
\setlength{\tabcolsep}{1mm}{
\begin{tabular}{|l|ccccc|}
\hline
\multirow{2}{*}{Method} & \multicolumn{1}{c|}{\multirow{2}{*}{MIT-Skating}} & \multicolumn{4}{c|}{Rhythmic Gymnastics}                                                    \\ \cline{3-6} 
                         & \multicolumn{1}{c|}{}                             & \multicolumn{1}{c|}{Ball} & \multicolumn{1}{c|}{Clubs} & \multicolumn{1}{c|}{Hoop} & Ribbon \\ \hline
Avg Pooling   & 0.605 & 0.518 & 0.650 & 0.650 & 0.552 \\
SAU           & 0.590 & 0.501 & 0.610 & 0.703 & 0.528 \\
LSTM + SAU    & 0.596 & 0.487 & 0.651 & 0.690 & 0.570 \\
Bi-LSTM + SAU & 0.572 & 0.522 & 0.568 & 0.674 & \textbf{0.600} \\ 
RTA \cite{doughty2019pros}			  & 0.611 & 0.522 & 0.634 & \textbf{0.713} & 0.565 \\ \hline
CAA~(ours)     & \textbf{0.615} & \textbf{0.528} & \textbf{0.657} & 0.708 & 0.578 \\ \hline
\end{tabular}}
\vspace{0.2cm}
\caption{Ablation study showing the contribution of the context-aware attention module in our method.
}
\label{tab:ablation stduy 2}
\vspace{-0.7cm}
\end{table}

\noindent{\textbf{Ablation Study of the Hybrid Dynamic-Static Architecture.}}
To show the effectiveness of the proposed hybrid dynamic-static architecture, we tried to remove either the dynamic stream or the static stream from the full model. The results of this ablation study are shown in Table \ref{tab:ablation study 1}. To save space, we use the abbreviations DS, SS, TS, and CAA to represent the dynamic stream only, the static stream only, the two streams together (the hybrid dynamic-static architecture) and our context-aware attention module, respectively. Interestingly, the results of SS + CAA show that methods using only the static stream can still achieve a good effect, even outperforming the methods using only the dynamic stream on the Hoop and Ribbon subsets of the Rhythmic Gymnastics dataset. These findings indicate that there is considerable redundant information in the dynamic stream and that the quality of actions performed by athletes can already be roughly assessed from only a few video frames. However, combining the dynamic stream with the static stream can boost the performance, as shown by the results of our full model (i.e., TS + CAA). 

\vspace{5pt}

\noindent{\textbf{Ablation Study of the Context-Aware Attention Module.}}
We also constructed four baseline methods by replacing the context-aware attention module in our method to evaluate its effectiveness:
\begin{itemize}
\item  Avg Pooling: We used average pooling at the temporal level to generate video-level descriptions. With this approach, the temporal evolution and timing of an action is lost.
\item SAU: We also used a standard attention unit (SAU) without context features as the input to evaluate the effectiveness of the context-aware attention module. The SAU was the same as the ATT-U used in the context-aware attention module.
\item RTA: We also compared our approach with multi-filter attention module from the  Rank-aware Temporal Attention (RTA) \cite{doughty2019pros} in our ablation study, in which the number of filters was set to three.
\item LSTM/Bi-LSTM + SAU: Because an LSTM architecture is often used for tasks with a sequential structure, we used an LSTM architecture to capture context information as the input to the SAU. We also used a Bi-LSTM architecture in place of the LSTM architecture. The hidden dimensions of the LSTM/Bi-LSTM layers were set to 256/128.
\end{itemize}

In Table \ref{tab:ablation stduy 2}, we compare the results of our method with those of the different baselines. Importantly, our proposed method shows improvements over all the alternative variants listed above on all datasets (i.e., our full method outperforms the Avg Pooling by 2.3 \% on the MIT-Skating dataset). When the context information captured by the TCG-U is removed, the performance drops considerably. Using an LSTM/Bi-LSTM module instead of the TCG-U to capture context information results in a decrease in model performance. By contrast, using average pooling alone also results in competitive performance, particularly for the Ball and Hoop subsets.
Similar to the experimental results of Doughty et al. \cite{doughty2019pros}, although the SAU achieves higher accuracy than average pooling for some tasks, we find the SAU results to be highly inconsistent for action quality assessment in long videos. 
Moreover, our context-aware attention slightly outperforms the multi-filter attention in the Rank-aware Temporal Attention \cite{doughty2019pros}, and the success of our context-aware attention and multi-filter attention indicate that the attention mechanism is an effective method for action assessment in long videos.

Additionally, to visualize the operation of the context-aware attention module, we present the attention weights computed for several segments of two videos in Figure \ref{fig:attention}. In Figure \ref{fig:visual}, we show further details of video clips with high and low attention weights from both videos; in this figure, each video segment is compressed to eight frames. We believe that if a particular video segment or frame has a high weight, then the video segment or frame shows an important technical movement that is likely to contribute to the final score; otherwise, it is not significant. With uniform weighting, the weight of each video segment is $1/28$~(0.0384), where the value of 28 corresponds to the number of segments per video used to train our network. Notably, due to the scoring rules of gymnastics and figure skating, it is difficult to accurately assess the quality of an athlete's actions by observing only a few video segments. Therefore, there are no order-of-magnitude differences in the weights between significant and insignificant segments. From Figure \ref{fig:visual}, we find that segments that show the athlete performing movements such as rolling or lifting one leg represent key technical movements and thus have very high attention values. From Figure \ref{fig:static stream}, we can also reach a similar conclusion in the static stream.

\vspace{5pt}

\noindent{\textbf{Is it necessary to perform human detection on the static \\ stream?}} In this paragraph, we discuss whether it is necessary to perform human detection on the static stream. To show the effectiveness of human detection on the static stream, we used features extracted from the whole scene as the input for our full model. As seen in Table \ref{tab:ablation study 3}, without human detection on the static stream, the performance achieved using either the static stream alone or the two streams combined is significantly decreased. Since athletes in sports videos often occupy only a small part of the image, it is difficult to extract high-quality spatial features of such an athlete when the entire image is used as the input.

\begin{table}[]
\setlength{\tabcolsep}{0.8mm}{
\begin{tabular}{|l|ccccc|}
\hline
\multirow{2}{*}{} & \multicolumn{1}{c|}{\multirow{2}{*}{MIT-Skating}} & \multicolumn{4}{c|}{Rhythmic Gymnastics}                                                      \\ \cline{3-6} 
                  & \multicolumn{1}{c|}{}                             & \multicolumn{1}{c|}{Ball} & \multicolumn{1}{c|}{Clubs} & \multicolumn{1}{c|}{Hoop} & Ribbon \\ \hline
$\textup{SS + CAA}^{\textup{w/o detection}}$    & 0.565 & 0.334 & 0.582 & 0.602 & 0.391 \\
$\textup{SS + CAA}^{\textup{w detection}}$      & 0.572 & 0.443 & 0.581 & 0.686 & 0.540 \\
$\textup{TS + CAA}^{\textup{w/o detection}}$    & 0.592 & 0.365 & 0.632 & 0.656 & 0.540 \\
$\textup{TS + CAA}^{\textup{w detection}}$      & \textbf{0.615} &\textbf{ 0.528} & \textbf{0.657} & \textbf{0.708} & \textbf{0.578} \\ \hline
\end{tabular}}
\vspace{0.2cm}
\caption{Ablation study showing the contribution of human detection in our method. 
}
\label{tab:ablation study 3}
\vspace{-1cm}
\end{table}

\vspace{-0.1cm}

\section{Conclusion}

In this work, we have proposed a hybrid dynamic-static context-aware attention network (ACTION-NET) for learning both video motion information and specific posture information from sampled video frames for action quality assessment. The context-aware attention module, which is applied to both streams in the proposed network, is able to learn useful representations by learning the relations between instances. 
The experimental results clearly show that our proposed ACTION-NET method can achieve state-of-the-art performance on two datasets.
Additionally, to support research on action quality assessment in long videos, we have collected and annotated the new Rhythmic Gymnastics dataset.

\vspace{-0.1cm}

\section{Acknowledgement}

This work was supported partially by the National Key Research and Development Program of China (2018YFB1004903), NSFC(U1911401, U1811461), Guangdong Province Science and Technology Innovation Leading Talents (2016TX03X157), Guangdong NSF Project (No. 2018B030312002), Guangzhou Research Project (201902010037), Research Projects of Zhejiang Lab (No. 2019KD0AB03), the Key-Area Research and Development Program of Guangzhou (202007030004), and Pearl River S\&T Nova Program of Guangzhou (201806010056).
We thank Jia-Chang Feng for useful feedback and suggestions.

\bibliographystyle{ACM-Reference-Format}
\bibliography{main}

\end{document}